\pdfoutput=1

\documentclass[11]{article}

\usepackage[]{acl}
\usepackage{times}
\usepackage{latexsym}

\usepackage[T1]{fontenc}

\usepackage[utf8]{inputenc}
\usepackage{pifont}
\newcommand{\cmark}{\ding{51}}%
\newcommand{\xmark}{\ding{55}}%
\usepackage{soul}
\usepackage{microtype}
\usepackage{tabularx}
\usepackage{xspace}
\usepackage{makecell}
\usepackage{tipa}
\usepackage{graphicx}
\usepackage{booktabs}
\usepackage{multirow}
\usepackage{caption}
\usepackage{amsmath} 
\usepackage{arydshln}
\usepackage{tcolorbox}
\usepackage{subfloat}
\usepackage{stmaryrd}
\usepackage{bm}
\usepackage{bbm}
\usepackage[normalem]{ulem}
\usepackage{amssymb}  
\usepackage{amsfonts}
\usepackage{multirow}
\usepackage{amssymb}
\usepackage{subfloat}
\usepackage{graphicx}
\usepackage{arydshln}
\usepackage{amsmath}
\usepackage{caption}
\usepackage{CJKutf8}
\usepackage{caption}
\usepackage{capt-of}
\usepackage{CJKutf8}
\usepackage{amsfonts}
\usepackage{pifont}
\usepackage{booktabs}
\usepackage{pifont}
\usepackage{stmaryrd}
\usepackage{wrapfig}
\usepackage{algorithm}
\usepackage{makecell}
\usepackage{tcolorbox}
\usepackage{bm}
\usepackage{bbm}
\usepackage{amssymb}
\usepackage{enumitem}
\usepackage{xcolor}
\usepackage[normalem]{ulem}
\usepackage{tabularx}
\usepackage{xspace}
\usepackage{subcaption}
\usepackage{footmisc}
\usepackage{marvosym}

\title{DEEM: Dynamic Experienced Expert Modeling for Stance Detection}

\author{Xiaolong Wang\textsuperscript{*,1}, Yile Wang\textsuperscript{*,2}, Sijie Cheng\textsuperscript{1,2}, Peng Li\textsuperscript{\Letter,2}, Yang Liu\textsuperscript{\Letter,1,2} \\
  \textsuperscript{1}Department of Computer Science and Technology, Tsinghua University, Beijing, China \\
  \textsuperscript{2}Institute for AI Industry Research (AIR), Tsinghua University, Beijing, China \\
    \texttt{wangxl22@mails.tsinghua.edu.cn, wangyile@air.tsinghua.edu.cn}\\
  \texttt{lipeng@air.tsinghua.edu.cn, liuyang2011@tsinghua.edu.cn}
  }

\DefineFNsymbols*{1}{*}
\setfnsymbol{1}

\begin{document}
\maketitle

\renewcommand{\thefootnote}{\fnsymbol{footnote}} 
    \footnotetext[1]{Equal contribution.}
\renewcommand{\thefootnote}{\arabic{footnote}}

\DefineFNsymbols*{1}{\Letter}
\setfnsymbol{1}

\renewcommand{\thefootnote}{\fnsymbol{footnote}} 
    \footnotetext[1]{Corresponding authors.}
\renewcommand{\thefootnote}{\arabic{footnote}}
\begin{abstract}
Recent work has made a preliminary attempt to use large language models (LLMs) to solve the stance detection task, showing promising results. However, considering that stance detection usually requires detailed background knowledge, the vanilla reasoning method may neglect the domain knowledge to make a professional and accurate analysis. Thus,
there is still room for improvement of LLMs reasoning, especially in leveraging the generation capability of LLMs to simulate specific experts (i.e., multi-agents) to detect the stance. In this paper, different from existing multi-agent works that require detailed descriptions and use fixed experts, we propose a Dynamic Experienced Expert Modeling (DEEM) method which can leverage the generated experienced experts and let LLMs reason in a semi-parametric way, making the experts more generalizable and reliable. Experimental results demonstrate that DEEM consistently achieves the best results on three standard benchmarks, outperforms methods with self-consistency reasoning, and reduces the bias of LLMs.

\end{abstract}

\section{Introduction}
Stance detection ~\citep{hasan-ng-2014-taking, kuccuk2020stance} is a natural language processing (NLP) task that automatically identifies the stance towards a specific target in a given text.
For example, the stance of ``{\it Secretary SecPompeo is as corrupt as every other member of the Trump}'' is \textit{against} Donald Trump.
Such a task has been shown to play an important role in gaining insights into public opinion ~\citep{public-opinion,LAI2020101075}, understanding political polarization ~\citep{JWS-0013}, and tracking ideological trends from social media~\citep{conforti-etal-2020-will}.

Recently, large language models (LLMs; \citealp{gpt3,palm,instructgpt, wang2023openchat}) are developing rapidly and can be applied to various tasks. For example,~\citet{zhang2022would} empirically confirms that ChatGPT can achieve impressive performance to detect stance in a zero-shot setting.~\citet{zhang2023investigating} further improve the results by using chain-of-thought reasoning strategies~\citep{cot, cheng2023unsupervised}. These works have opened up new directions in stance detection.

Despite the success of applying LLMs, conventional reasoning techniques with LLMs could cause hallucinations and factual errors~\citep{guerreiro-etal-2023-looking,ji2023survey}, particularly in stance detection.
Texts in stance detection usually originate from social media~\cite{aldayel2021stance}, which are typically short and intricate, necessitating additional domain expertise~\cite{he-etal-2022-infusing}.
For example, to detect the stance towards Biden in ``\textit{Are you actually trying, as president of the U.S., to start a war??!! \#VoteBlueToSaveAmerica2020 \#Biden}'', we need to know which camp Biden belongs to and the meaning of ``\#VoteBlueToSaveAmerica2020''. 
 

\begin{figure}[t!]
    \centering
    \includegraphics[width=\columnwidth]{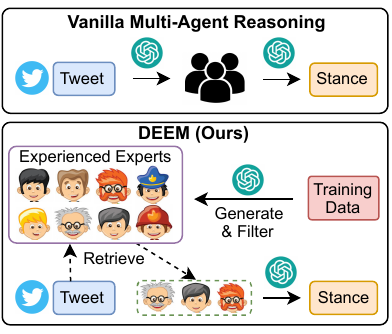}
    \caption{Illustration of our DEEM method. Top: Vanilla multi-agent reasoning for stance detection through generation. Bottom: Our method first generates and filters experienced experts by leveraging training data, then retrieves the related ones during reasoning.}
    \label{fig:intro}
\end{figure}

Inspired by \textit{the wisdom of crowds} in sociological theory~\citep{minsky1988society, piaget2013construction}, we intuitively propose designing multiple capable experts to collaborate in order to come up with a comprehensive stance prediction. 
Previous studies~\citep{du2023improving, wang2023unleashing} have attempted to solve reasoning tasks with multi-agent debate and multi-persona self-collaboration.
However, their designed agents are generally pre-defined or automatically generated by LLMs, which either require strong prior knowledge or need to be further improved for stance detection tasks.
Obviously, pre-defined agents are fixed, thus it is difficult to adapt to different contexts in social media.
Moreover, fully generated agents by LLMs may not be suitable due to the intricate contextualized information, especially in specific domains.



In this work, we propose DEEM, a \underline{D}ynamic \underline{E}xperienced \underline{E}xpert  \underline{M}odeling method to solve stance detection tasks, as shown in Figure~\ref{fig:intro}.
In particular, to better gather the potential expertise for stance detection, we first leverage labeled samples from the existing training data to generate diverse experts.
Then, we design two heuristic rules, namely occurrence numbers and response accuracy, to filter the experienced experts and construct an expert repository.
Finally, instead of using a fully generative approach, we adopt a dynamic retrieval method to identify relevant experienced experts for new sentences, facilitating discussions for the final prediction. 

We evaluate DEEM across both single-target and multi-target stance detection tasks on three widely used datasets, including P-Stance~\citep{pstance}, SemEval-2016~\citep{semeval2016}, and  MTSD~\citep{mtsd}.
Experimental results demonstrate that DEEM with dynamic experienced experts can gain substantial improvement across all datasets.
Furthermore, it also outperforms reasoning with self-consistency that requires multiple responses and shows potential for reducing the bias of LLMs. Code is available at \url{https://github.com/THUNLP-MT/DEEM}.

\section{Related Work}
\textbf{Stance Detection}. Early works on stance detection mainly take it as a classification task, leveraging the small language models~\citep{bert,bertweet} and learning features from either in-domain or cross-domain training datasets~\citep{bicond,zhang-etal-2019-aspect, toad,liu-etal-2021-enhancing,liang-etal-2022-jointcl}. 
With the emergence of LLMs, \citet{zhang2022would,zhang2023investigating} first try using ChatGPT to solve the task directly by zero-shot or few-shot reasoning with chain-of-thought, which only requires simple prompts to obtain the political stance from the generated responses. In comparison to their methods, we take inspiration from the multi-agent~\citep{wang2023survey,xi2023rise} and introduce a novel dynamic expert mechanism, enabling LLMs to generate responses from multiple perspectives, providing more comprehensive responses and improve the prediction accuracy.

\noindent\textbf{LLMs Reasoning with Multi-Agent}. Multi-agent strategies have proven to be effective in LLMs reasoning~\cite {talebirad2023multi,li2023camel}. By using prompts or instructions that specify the desired role or persona, the model can generate responses based on its understanding of that role and can apply to more complicated scenarios such as social interaction~\citep{park2023generative}, court simulation~\citep{hamilton2023blind}, code development~\citep{qian2023communicative}, and engaging communication games~\citep{xu2023exploring}.
The above works often require detailed role specialization at the beginning of each task~\citep{xu2023expertprompting}. In contrast, we propose to discover the experienced personas automatically with the least human effort by using the existing labeled samples, then call for dynamic expertise by retrieving the collected ``sentence-expert'' pairs. 
The entire process minimizes the prior knowledge required from stance detection.

\section{Methods}

\begin{figure}[t!]
	\centering
	\includegraphics[width=\columnwidth]{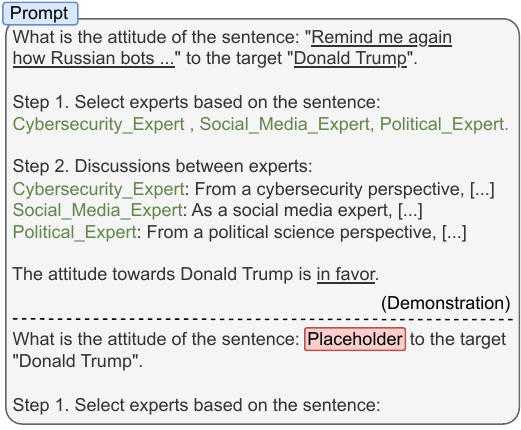}
	\caption{An example of a few-shot prompt with expert modeling for stance detection. The underlined parts are the sentence, target, and label, respectively. The green texts indicate the manually written experts.}
	\label{figure:prompt}
\end{figure}

\begin{figure*}[t!]
	\centering
	\includegraphics[width=\linewidth]{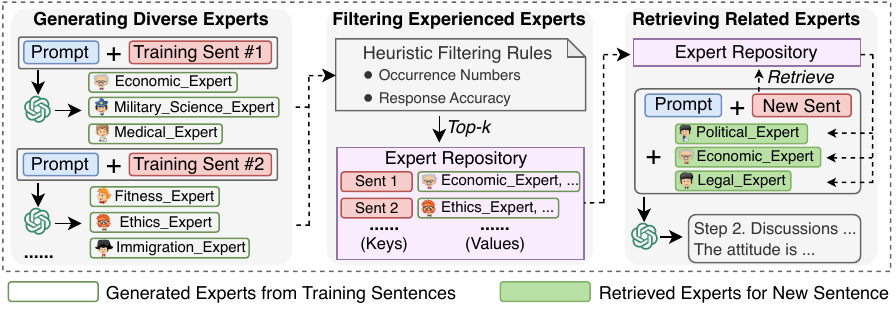}
	\caption{The overall pipeline of our DEEM method. We first use the prompt and training sentences to generate diverse experts ($\S$~\ref{section:31}). Then we filter experienced experts according to their occurrence numbers and performance ($\S$~\ref{section:32}). Finally, we build the sentence-expert pairs and retrieve the experienced experts for each new sentence ($\S$~\ref{section:33}).}
	\label{figure:pipeline}
\end{figure*}


The overall pipeline of the proposed DEEM is shown in Figure~\ref{figure:pipeline}. DEEM first generates diverse experts by leveraging the training data. Then it filters the experienced experts that are generalizable and reliable. Finally, it retrieves the experts according to new queried sentences and makes responses.

\subsection{Generating Diverse Experts}
\label{section:31}
Conventionally, training datasets are used to fine-tune the parameters of small models. Recently, they can be chosen as prompts for better few-shot in-context learning for LLMs~\cite{gpt3-icl,auto-cot,shum2023automatic}. In this paper, we leverage the existing training datasets in a novel way to help generate potentially useful experts for solving stance detection, without using much prior knowledge or detailed role descriptions.

Firstly, given the training dataset with sentence-target-label triplets $\mathcal{D}=\{s_j,t_j,l_j\}_{j=1}^ {|\mathcal{D}|}$, we randomly select some of them as held-out ones to construct a prompt as shown in Figure~\ref{figure:prompt}:
\begin{equation}
    {\rm prompt} = s_p \oplus t_p \oplus \mathcal{E}_p \oplus l_p,\
\end{equation}
where $\oplus$ denotes textual concatenation, $s_p$, $t_p$, $l_p$ are the sentence, target, and label of the selected instance in prompt, $\mathcal{E}_p = \{e^1_p, ..., e^k_p\}$ are the \textit{manually written} experts corresponding to the selected instance, such as ``Social Media Expert''.

Then, we use the LLM $\mathcal{M}$ to generate experts and predicted labels for all other sentences via few-shot in-context learning~\citep{gpt3}:
\begin{equation}
    \mathcal{E}_j,\hat{l}_j = {\mathcal{M}} ( {\rm prompt} \oplus s_j \oplus t_j),
    \label{eq:1}
\end{equation}
where $\mathcal{E}_j = \{e^1_j$, ..., $e^k_j\}$ denotes the \textit{generated} experts, and $\hat{l}_j$ indicates the predicted label to the sentence $s_j$ and target $t_j$.

It is worth noting that more than 1,400 distinct experts can be generated, showing LLMs' strong capability of in-context learning.
We further show the detailed expert distributions in Section~\ref{section:expert_distribution}.

\subsection{Filtering Experienced Experts}
\label{section:32}
The generated expert candidates in $\mathcal{E}$ are directly generated by LLMs, which are diverse enough but many do not always match the sentences. To make them more generalizable and reliable to new sentences, we designed two heuristic rules to filter the experienced experts among all candidates.

First, despite a large number of generated experts, many of them only appear a few times. These low-frequency experts are usually from unrelated domains, making it difficult to generalize to new sentences for stance detection.
Thus, we require experts who are experienced in stance detection tasks.
To fulfill this requirement, the first heuristic rule is the total occurrence numbers ${\rm Count}(\cdot)$ of each expert $e_j^m \in \mathcal{E}_j$ in the training dataset as below:
\begin{equation}
{\rm Count}(e_j^m) = \sum_{e_i \in \mathcal{E}} \mathbbm{1}({e_j^m = e_i)},
\end{equation}
where $e_i$ is the $i$-th element in the collection of generated experts $\mathcal{E}$, and $\mathbbm{1}(e_j^m = e_i)$ is the indicator function that equals 1 when $e_j^m$ equals to $e_i$ and 0 otherwise.

Second, it is well known that LLMs can occasionally generate hallucinations~\cite{guerreiro-etal-2023-looking,ji2023survey}, thus they could make the responses unreliable and lead to incorrect final predictions.
Therefore, experienced experts need to be accurate in analyzing the stance towards the target, thus the second heuristic rule is the total prediction accuracy ${\rm Acc} (\cdot)$ of each expert $e_j^m \in \mathcal{E}_j$:
\begin{equation}
\begin{aligned}
{\rm Acc} (e_j^m) = \frac{\sum_{e_j^m = e_i} \mathbbm{1}(\hat{l}_i = l_i) }{\sum_{e_i \in \mathcal{E}} \mathbbm{1}({e_j^m = e_i)}},
\end{aligned}
\end{equation}
where $\hat{l}_i$ is the predicted label corresponding to the $i$-th element $e_i$ in Eq.~\ref{eq:1}, and $l_i$ is the ground-truth label for sentence $s_i$.

Finally, we discard the expert $e_j^m \in \mathcal{E}_j$ who have low prediction accuracy as the threshold (e.g., ${\rm Acc} (e^m_j) < 50\%$).
Then we select the rest of the top-$k$ experts (e.g., $k=10\sim30$) according to their occurrence numbers ${\rm Count}(e_j^m)$.
We take these selected experts $\mathcal{E}'_j$ in each sentence $s_j$ as the final experienced experts.
Moreover, we regard all of them as the expert pool $\mathcal{E}' = \mathcal{E}'_1 \cup \cdots \cup \mathcal{E}'_{|\mathcal{D}|}$ to solve the stance detection for a new sentence. We discuss the settings of ${\rm Acc} (e^m_j)$ and top-$k$ in Section~\ref{section:filtering_strategy}.

\subsection{Retrieving Related Experts}
\label{section:33}

Verified on the training set, all the experts in the expert pool are experienced with both high occurrence and accuracy, these pieces of information can be utilized during the testing phase for retrieval and placed in the prompt as inputs to the model~\cite{guo2023prompt,wang-etal-2023-self-knowledge}. Overall, to ascertain the stance of a new sentence, it is more effective to choose experienced experts from similar sentences, rather than having LLMs generate potentially unrelated or inexperienced experts directly.

Specifically, we construct a repository to better match the new sentence and these experienced experts.
We assign the sentence $s_j$ in the training dataset as the key.
Meanwhile, its corresponding filtered experienced experts $\mathcal{E}'_j$ as the value.
The resulting sentence-expert pair $\langle s_j,\mathcal{E}'_j\rangle$ indicates that the filtered experienced experts $\mathcal{E}'_j$ can accurately and expertly detect the stance of the sentence $s_j$, thus they have potential to be applied to the similar sentence to $s_j$.
 
In the inference phase, given a new sentence $s$ and the constructed sentence-expert repository, we can retrieve the top-$h$ related experienced experts according to the textual similarity scores: 
\begin{equation}
\begin{aligned}
    \text{Sim}(s, s_j) = \frac{\exp({\rm Enc}(s)\cdot  {\rm Enc}(s_j))}{\sum_{i=1}^{\left|\mathcal{D}\right|}\exp({\rm Enc}(s)\cdot {\rm Enc}(s_i))},
\end{aligned}
\end{equation}
where ${\rm Enc}(\cdot)$ is a sentence encoder, such as SimCSE~\cite{gao-etal-2021-simcse}.

\begin{table}[t!]
\small
    \centering
    \resizebox{\columnwidth}{!}{
	\begin{tabular}{lcccc}\toprule
		
		\multirow{2}*{\textbf{Method}}&\textbf{Including}&\textbf{Multi-} & \textbf{Verified} &\textbf{Reasoning} \\
  &\textbf{Explanations}&\textbf{Roles} & \textbf{Experts} &\textbf{Type} \\
        \midrule
        Few-Shot&\textcolor{red}{\xmark} & \textcolor{red}{\xmark}& -   & Gen \\
		CoT&\textcolor{teal}{\cmark}& \textcolor{red}{\xmark} & -  & Gen\\
        Auto-CoT&\textcolor{teal}{\cmark}& \textcolor{red}{\xmark} & -  & Re+Gen\\
        ExpertPrompt& \textcolor{teal}{\cmark}& \textcolor{red}{\xmark} & \textcolor{red}{\xmark}  & Gen\\
		SPP&\textcolor{teal}{\cmark}&  \textcolor{teal}{\cmark}& \textcolor{red}{\xmark } & Gen\\

        DEEM (ours)& \textcolor{teal}{\cmark}& \textcolor{teal}{\cmark} & \textcolor{teal}{\cmark}  & Re+Gen\\
		\bottomrule
	\end{tabular}}
	\caption{Comparison to typical reasoning methods including Few-Shot, chain-of-thought (CoT), Auto-CoT, ExpertPrompt, and solo performance prompting (SPP). Re: Retrieval. Gen: Generation.}
	\label{table:method_comparison}
\end{table}

Finally, we obtain the top-$h$ experienced experts and directly append them to the prompt as shown in Figure~\ref{figure:pipeline}. 
Then we use the whole prompt as the input for LLMs to generate the upcoming experts' discussion and the final predicted answer.

\subsection{Comparison with Other Methods}
We compare our method with typical reasoning approaches in Table~\ref{table:method_comparison}. To involve explanations or specific roles during the reasoning process, CoT~\cite{cot}, ExpertPrompt~\cite{xu2023expertprompting} and SPP~\cite{wang2023unleashing} let LLMs fully generate the explanations or discussions between experts by using the prompt. As for our method, we propose to explore experienced experts from training samples and introduce a retrieval mechanism during the reasoning process. 

According to the retrieval mechanism~\citep{borgeaud2022improving}, Auto-CoT~\cite{auto-cot} finds similar samples as demonstrations \textit{in the prompt}. In contrast, we retrieve according to constructed ``sentence-expert'' pairs \textit{during reasoning}, which makes the involved experts more related to the current sentence in a semi-parametric manner.

\section{Experiments}
\subsection{Experimental Setups}
\textbf{Datasets.}
To evaluate the effectiveness of our method, we comprehensively use three standard stance detection datasets, including both single-target and multi-target tasks: (1) P-stance~\citep{pstance} is a political stance detection dataset, with each tweet annotated for its stance towards one of three politicians;
(2) SemEval-2016~\citep{semeval2016}  introduces a shared task on stance detection from tweets, including six targets with one target exclusively for testing;
(3) MSTD~\citep{mtsd} is a dataset for multi-target stance detection, primarily focusing on four presidential candidates in the 2016 US election using specific hashtags.
Among these datasets, we mainly focus on politicians, such as ``Donald Trump''.
The detailed statistics are shown in Table~\ref{table:statistics}.

\begin{table}[t!]
\small
    \centering
    \resizebox{0.98\columnwidth}{!}{
	\begin{tabular}{lccc}\toprule
		
		\textbf{Datasets}&\textbf{Target}&\textbf{Train} & \textbf{Test} \\
        \midrule
		\multirow{3}*{P-Stance}&Trump &  6,362 &796\\
		&Biden &  5,806 &745\\
		&Sanders &  5,056 &635\\
            \midrule
		\multirow{2}*{SemEval-2016}&Clinton &  1,898 &984\\
		&Trump &  2,194 &707\\
            \midrule
		\multirow{3}*{MTSD}&Trump-Clinton& 1,240 &355\\
            &Trump-Cruz& \ \ \ 922  &263\\
            &Clinton-Sanders& \ \ \ 957 &272\\
		\bottomrule
	\end{tabular}}
	\caption{Statistics of P-Stance, SemEval-2016, and MTSD datasets in our experiments.}
	\label{table:statistics}
\end{table}

\begin{table*}[t!]
\small
	\centering
    \resizebox{\linewidth}{!}{
	\begin{tabular}{clccccccccc}
	    \toprule
        \multirow{2.5}*{\textbf{Type}}&\multirow{2.5}*{\textbf{Method}}&\multicolumn{3}{c}{\textbf{P-Stance}}&\multicolumn{2}{c}{\textbf{SemEval-2016}}&\multicolumn{3}{c}{\textbf{MTSD}}&\multirow{2}*{\textbf{Avg.}}\\
        \cmidrule(lr){3-5}\cmidrule(lr){6-7}\cmidrule(lr){8-10}
        &&DT&JB&BS&HC&DT&DT-HC&DT-TC&HC-BS&\\
    	\midrule
    \multirow{4}*{FT}
     &BiCond~\citep{bicond}$^\spadesuit$&73.0&69.4&64.6&\ 32.7$^\dag$&\ 30.5$^\dag$&-&-&-&-\\
     &BERT~\citep{bert}$^\spadesuit$&81.6&81.7&78.4&\ 49.6$^\dag$&\ 40.1$^\dag$&-&-&-&-\\
     &BERTweet~\citep{bertweet}&82.4&81.0&78.1&\ 50.9$^\dag$&\ 42.2$^\dag$&69.2&70.7&69.0&67.9\\
     &JointCL~\citep{liang-etal-2022-jointcl}$^\heartsuit$ &-&-&-&\ 54.8$^\dag$&\ 50.5$^\dag$&-&-&-&-\\

     \midrule
     \midrule
     \multicolumn{11}{c}{ (\texttt{text-davinci-003})} \\
     \addlinespace[2pt]
             \multirow{2}*{ZS}&Zero-Shot~\citep{gpt3}&73.8&83.3&77.5&71.8&68.3&61.6&64.7&61.4&70.3\\
    &DQA~\citep{zhang2022would}&73.0&80.8&76.1&72.7&69.9&58.9&66.4&63.3&70.1\\
    \addlinespace[1pt]
     \cdashline{1-11}
     \addlinespace[3pt]
     \multirow{7}*{FS}&Few-Shot~\citep{gpt3}&79.9&85.2&78.6&79.4&73.5&68.6&65.9&70.7&75.2\\
    
     \multirow{7}*{($d=2$)}&Manual-CoT~\citep{cot}&79.3&84.9&78.4&77.2&72.5&\underline{75.0}&75.6&68.8&76.5\\
     &StSQA~\citep{zhang2023investigating}&75.2&85.2&78.9&78.3&72.3&72.6&75.9&72.0&76.3\\
     &Auto-CoT~\citep{auto-cot}&82.9&84.7&78.4&80.7&\underline{73.8}&67.9&67.4&75.3&76.4\\
     &ExpertPrompt~\citep{xu2023expertprompting}&82.8&\underline{85.5}&78.7&85.2&73.0&74.1&76.8&71.5&78.5\\
     &SPP~\citep{wang2023unleashing}&\underline{83.4}&\underline{85.5}&\underline{79.6}&\underline{85.5}&73.3&73.0&\underline{78.0}&\underline{76.8}&\underline{79.4}\\
     &\textbf{DEEM} (ours)&\textbf{83.7}&\textbf{86.0}&\textbf{80.4}&\textbf{85.7}&\textbf{74.8}&\textbf{76.5}&\textbf{80.1}&\textbf{81.3}&\textbf{81.1}\\
     &$\Delta$\ (\text{compare w/ second-best results})&\textcolor{teal}{+0.3} &\textcolor{teal}{+0.5} &\textcolor{teal}{+0.8} & \textcolor{teal}{+0.2}& \textcolor{teal}{+1.0}&\textcolor{teal}{+1.5} & \textcolor{teal}{+2.1}& \textcolor{teal}{+4.5}& \textcolor{teal}{+1.7}\\
     \midrule
     \midrule
     \multicolumn{11}{c}{ (\texttt{gpt-3.5-turbo-0301})} \\
     \addlinespace[2pt]
     \multirow{2}*{ZS}&Zero-Shot~\citep{gpt3} &83.3&82.5&79.4&79.3&71.4&73.5&67.0&73.6&76.3\\
&DQA~\citep{zhang2022would}$^\spadesuit$&83.2&82.0&79.4&78.0&71.3&66.2&63.2&69.3&74.1\\
\addlinespace[1pt]
\cdashline{1-11}
\addlinespace[3pt]
     \multirow{7}*{FS}&Few-Shot~\citep{gpt3} &83.6&83.1&80.8& 79.3& 71.6&76.6&78.2&72.8&78.3\\
     \multirow{7}*{($d=2$)}&Manual-CoT~\citep{cot}&85.4&83.8&80.9&79.5&71.2&77.0&77.5&76.7&79.0\\
     &StSQA~\citep{zhang2023investigating}$^\spadesuit$&\underline{85.7}&82.8&80.8&78.9&71.6&77.5&78.2&\underline{81.2}&79.6\\
     &Auto-CoT~\citep{auto-cot}&84.1&82.8&80.6&84.6&73.5&77.0&76.9&76.7&79.5\\
     &ExpertPrompt~\citep{xu2023expertprompting}&84.7&\underline{84.7}&81.2&83.8&77.4&\underline{80.6}&77.0&79.0&81.1\\
     &SPP~\citep{wang2023unleashing}&85.1&84.6&\underline{81.5}&\underline{85.3}&\underline{79.5}&79.5&\underline{79.8}&79.8&\underline{81.9}\\
       &\textbf{DEEM} (ours) &\textbf{86.4}&\textbf{86.1}&\textbf{82.1}&\textbf{85.9}&\textbf{80.5}&\textbf{81.7}&\textbf{80.7}&\textbf{83.5}&\textbf{83.4}\\
 &$\Delta$\ (\text{compare w/ second-best results})&\textcolor{teal}{+0.7} &\textcolor{teal}{+1.4} &\textcolor{teal}{+0.6} & \textcolor{teal}{+0.6}& \textcolor{teal}{+1.0}&\textcolor{teal}{+1.1} & \textcolor{teal}{+0.9}& \textcolor{teal}{+2.3}& \textcolor{teal}{+2.5}\\
        
     \bottomrule
	\end{tabular}}

	\caption{Main results of baselines and our proposed DEEM. FT: Fine-tuning, ZS: Zero-shot, FS: Few-shot. DT:  Donald Trump, JB: Joe Biden, BS: Bernie Sanders, HC: Hillary Clinton, TC: Ted Cruz. $\dag$: cross-target setting. $\spadesuit$: reported by~\citet{zhang2023investigating}, $\heartsuit$: reported by~\citet{liang-etal-2022-jointcl}. The other results are achieved via our implementation. The best results are in \textbf{bold} and the second-best results are \underline{underlined}.}
	\label{table:mainresults}
\end{table*}

\noindent\textbf{Baselines.}
Besides the methods that require supervised fine-tuning, we also compared our method with recent methods by using LLMs without modifying model parameters:
\begin{itemize}
    \setlength\itemsep{-0.3em}
    \item DQA~\citep{zhang2022would} uses the template ``\textit{What is the attitude of the sentence :} \texttt{[Tweet]} \textit{to the target:} \texttt{[Target]}. `\textit{favor}' \textit{or} `\textit{against}'.'' and extract the answers by question answering. 
    \item CoT~\citep{cot} manually provides the explanations in demonstrations and enhances the chain-of-thought reasoning ability of LLMs. 
    \item StSQA~\citep{zhang2023investigating} proposes automatic ``thought-inducing'' and add them to the demonstrations for step-by-step question answering.
    \item Auto-CoT~\citep{auto-cot} automatically selects demonstrations from training data according to semantic diversity.
    \item ExpertPrompt~\citep{xu2023expertprompting} introduces the identity of experts and customizes information descriptions for LLMs before giving responses.
    \item SPP~\citep{wang2023unleashing} proposes solo performance prompting by engaging in multi-turn collaboration with multi-persona during reasoning.
\end{itemize}

\noindent\textbf{Metric.} Following~\citet{pt-hcl} and~\citet{zhang2023investigating}, we use the F1$_{avg}$, i.e., the average of F1-score on the label `\textit{favor}' and `\textit{against}', as the metric for evaluation and comparison.

\noindent\textbf{Implementation Details.} We employ InstructGPT (\texttt{text-davinci-003}) and ChatGPT (\texttt{gpt-3.5-turbo-0301}) as in \citet{zhang2022would, zhang2023investigating} through OpenAI API\footnote{\url{https://platform.openai.com/}, we also tried the newer version model of \texttt{gpt-3.5-turbo-0613}, but the results did not show significant improvement.}.
We set the number of demonstrations $d$ in both the first and third stages as 2. The number of experts (i.e., the number $k$) in the first stage is 3, and the number of top-$h$ in the third stage is also 3.
The temperature is set to 0 to ensure the reproducibility of the LLMs' responses.

\subsection{Main Results}

The main comparison results on three standard datasets are reported in Table~\ref{table:mainresults}.
For fine-tuning models, BERT and BERTweet achieve comparable results on P-Stance, while BERTweet achieves relatively good performance on both SemEval-2016 and MTSD datasets. Moreover, JointCL obtains the highest performance on SemEval-2016.
These phenomena show that small models can capture domain knowledge by fine-tuning useful data to further enhance the results.

\begin{table*}[t!]
\small
	\centering
{\resizebox{\linewidth}{!}{
	\begin{tabular}{ccccccc}
            
     \toprule
\multirow{2}*{\textbf{Target}}&\multicolumn{3}{c}{\textbf{Frequency (Proportion)}}&\multicolumn{3}{c}{\textbf{Accuracy (Proportion)}}\\   
\cmidrule(lr){2-4}\cmidrule(lr){5-7}
&\textbf{>1\% (2.98\%)}&\textbf{0.05-1\% (24.32\%)} &\textbf{<0.05\% (72.70\%)}&\textbf{>80\% (58.06\%)}&\textbf{50-80\% (10.67\%)} &\textbf{<50\% (31.27\%)}\\
\midrule

\multirow{3}*{\textbf{Sanders}}&Political&LeaderShip&Future\_Prediction&Immigration&Media&Banking\\

&Ethics &History&Progmatism &Economic&Political&Comedy\\
 
&Economic&Technology&Transparency&History&Polling&Alcohol\\
 
\midrule

\multirow{3}*{\textbf{Trump}}&Immigration&Corruption&Energy&Religious&Political&Slang\\

&Political &Social\_Policy&Ethanol &Election&Social\_Media&Nationality\\
 
&Military\_Science&Taxation&Deception&Ethics&Economic&Endorsement\\
 
\midrule

\multirow{3}*{\textbf{Clinton}}&Gender&Technology&Fash\_Food&Ethics&Unity&Geography\\

&Healthcare &Labor&Alcohol\_Policy&Legal&Gender&Diversity\\
 
&Political&Fashion&Leadership&National\_Security&Political&Anarchist\\
\midrule
\multirow{3}*{\textbf{Cruz}}&Media&Unity&Values&National\_Security&Political&Music\\

&Political &Technology&Nationality &History&Language&TeaParty\\
 
&Legal&Polling&Chess&Religious\_Studies&Election&Voting\\
 
    \bottomrule
	\end{tabular}}}
	\caption{Distributions and examples of diverse generated experts with different frequencies (Left) and prediction accuracy (Right) to different targets.}
	\label{table:all-experts-distribution}
\end{table*}


As for LLMs, zero-shot methods using InstructGPT can not surpass fine-tuning models in both the P-Stance and MTSD datasets, showing that LLMs with a larger number of parameters do not achieve better results without using specific strategies. 
Traditional few-shot learning and their reasoning methods obtain significant improvements, especially under InstructGPT, indicating demonstrations and chain-of-thoughts strategies are effective for solving complex tasks.
Moreover, the strong performances of ExpertPrompt and SPP prove that using experts is quite useful for stance detection.
However, their improvement is not always stable. For example, SPP performs well on P-stance tasks but struggles on SemEval-2016 by using Instruct.

Our method DEEM consistently yields superior results across all three datasets, regardless of whether the InstructGPT or ChatGPT model is utilized. One of the reasons for this superior performance is that DEEM effectively leverages the knowledge within LLMs and adapts it to specific tasks using expert domain knowledge. In some cases, the advantage of DEEM is particularly large (e.g., in the multi-target setting), possibly because it can better capture the underlying structure and relationships within the data. 
Moreover, compared with expert-based methods, the performance of DEEM is much more stable, indicating that introducing the retrieving module can recall more suitable experts.
Overall, DEEM can generalize well and perform better than all other methods, benefiting from dynamic experienced experts and our retrieval mechanism.




\section{Analyses and Discussion}
In this section, we conduct a series of analyses to probe the reason behind the effectiveness of our proposed method DEEM. Specifically, we mainly conduct these experiments on the only multi-target stance detection task, i.e., the MTSD dataset.

\subsection{Expert Distributions}
\label{section:expert_distribution}

We first investigate expert types according to their frequency and accuracy in the generating stage (Section~\ref{section:31}).
The examples and the distributions of the generated experts are shown in Table~\ref{table:all-experts-distribution}.

\noindent\textbf{Frequency.} As for the frequency, we can see that ``political experts'' appear many times for all targets, showing that LLMs can uncover the shared characteristics of political character and use this to solve the stance detection task. We also find that different targets can exhibit some distinctive types of experts. For example, the ``Gender Expert'' appears when the target is ``Hillary Clinton''. Overall, it shows an unbalanced distribution where 72.70\% experts have a low frequency, i.e., less than 0.05\%. Intuitively, the majority of these low-frequency experts (e.g., ``Ethanol Expert'', ``Chess Expert'') do indeed have limited generalizability for stance detection tasks. 

\noindent\textbf{Accuracy.} According to the accuracy, some experts do not show good performance for stance detection. For example, over 30\% experts achieve an accuracy of less than 50\%. As for the ``political experts'', the accuracy is mediocre (ranging from 50\% to 80\%), showing that high frequency does not always lead to high accuracy. Thus we need to combine both the frequency and accuracy to filter experienced ones for better solving stance detection tasks.

\subsection{Filtering Strategies}
\label{section:filtering_strategy}
\begin{figure}[thbp!]
    \centering
    \begin{subfigure}[b]{\columnwidth}
        \includegraphics[width=\textwidth]{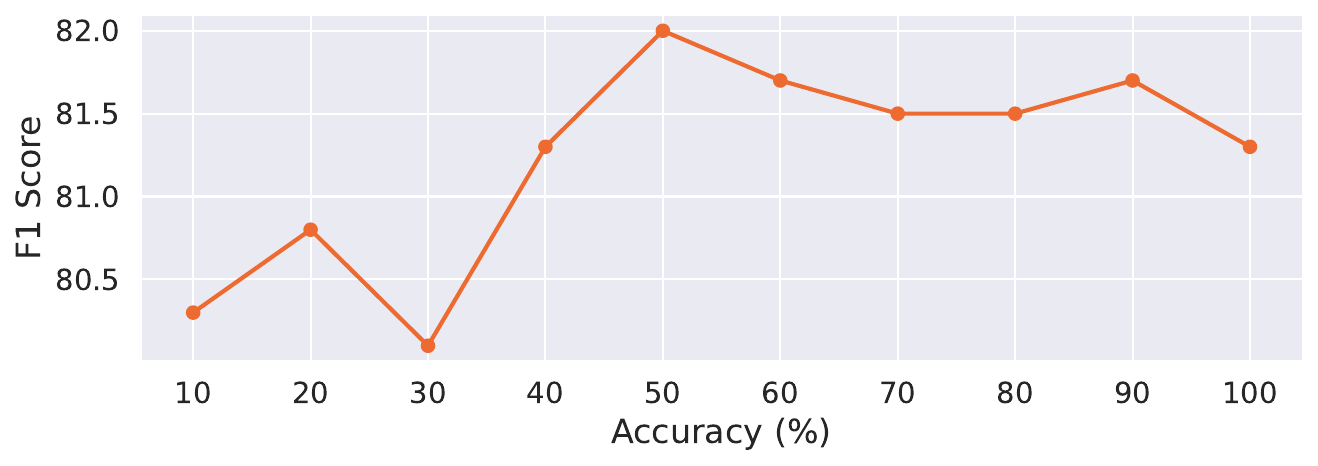}
        \label{fig:rate_gate}
    \end{subfigure}%
    \hfill
    \begin{subfigure}[b]{\columnwidth}
        \includegraphics[width=\textwidth]{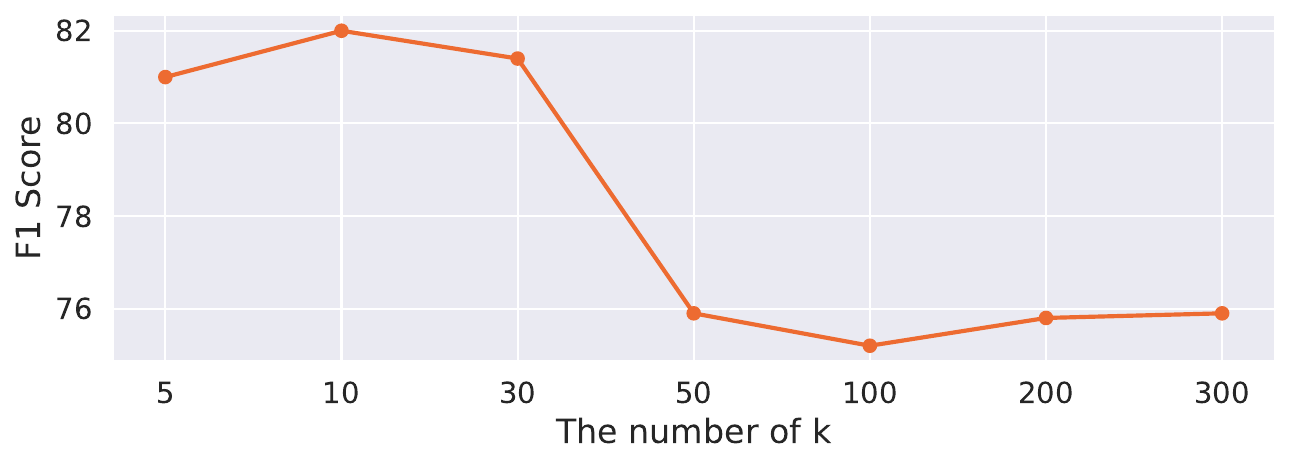}
        \label{fig:pool_size}
    \end{subfigure}%
    \caption{Impact of filtering strategies according to accuracy (Top) and frequency (Bottom).}
    \label{fig:filtering_strategies}
\end{figure}

We then investigate the accuracy and frequency threshold for filtering experienced experts (Section~\ref{section:32}), the averaged results for three multi-target pairs are shown in Figure~\ref{fig:filtering_strategies}.

\begin{figure*}[t!]
    \centering
    \includegraphics[width=\textwidth]{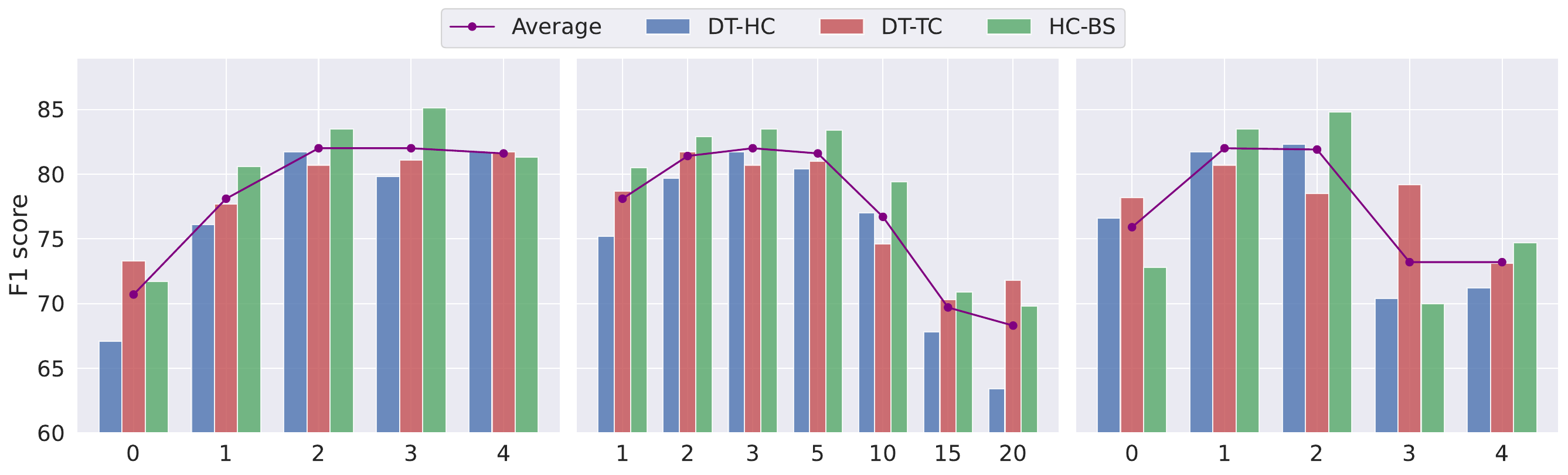}
    \par
    \begin{tabular}{>{\centering\arraybackslash}p{0.35\textwidth}>{\centering\arraybackslash}p{0.25\textwidth}>{\centering\arraybackslash}p{0.33\textwidth}}
        {\small (a) Number of Demonstrations} &
        {\small (b) Number of Experts} &
        {\small (c) Number of Discussion Turns}
    \end{tabular}
    \caption{Impact of (a) demonstrations, (b) retrieved experts, and (c) discussion turns during reasoning.}
    \label{fig:numbers}
\end{figure*}

\noindent\textbf{Accuracy Threshold.} 
The impact of the accuracy threshold is shown at the top of Figure~\ref{fig:filtering_strategies}. The results are relatively bad when the threshold is low (e.g., around 80\% when accuracy is lower than 30\%). This shows that the experts with low accuracy may not be generalizable enough. We find that the 50\% threshold leads to the best results (around 82\%), and the performance does not improve when the threshold continues to increase. The reason can be that an intermediate threshold can maintain both the generalization and diversity of potential useful experts for new sentences, thus showing the best results for test sentences.

\noindent\textbf{Frequency Threshold.}
We set different frequency thresholds, i.e., different top-$k$ selected experts in the expert pool. The results are shown at the bottom of Figure~\ref{fig:filtering_strategies}. We can see that 10$\sim$30 experts achieve better results. The performance largely reduces when the number of $k$ increases, showing the negative influence of involving the possible unrelated experts with useless information.

\subsection{Dynamic Experts vs. Fixed Experts}

\begin{figure}[t!]
    \centering
    \includegraphics[width=\columnwidth]{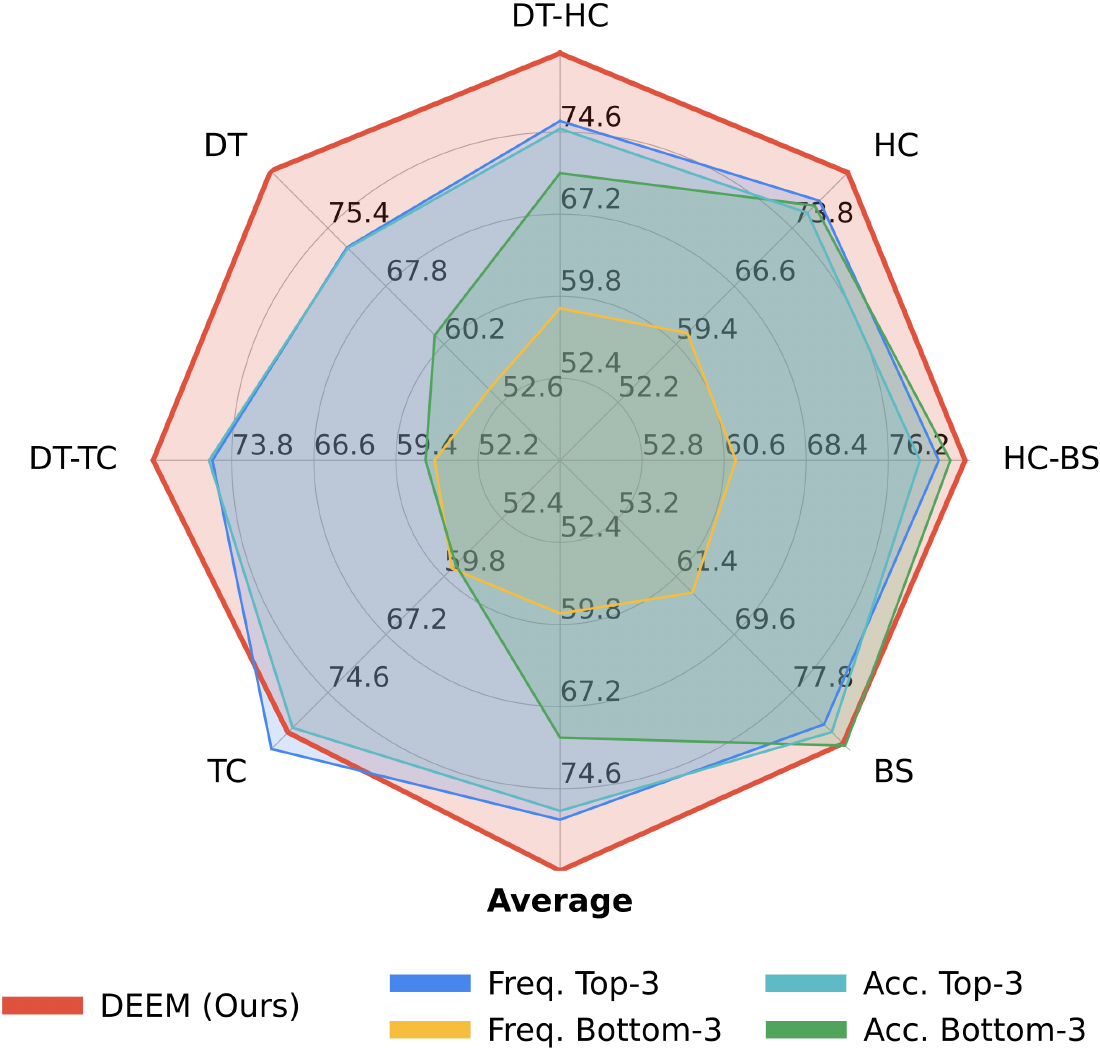}
    \caption{Comparing dynamic experts with fixed experts according to frequency and accuracy.}
    \label{fig:dynamic}
\end{figure}

To prove the effectiveness of our dynamic experts mechanism, we compare fixed experts that directly rely on frequency and accuracy for selecting experts. 
Specifically, we attempt to use the top-3 and bottom-3 experts according to frequency and accuracy. The results are shown in Figure~\ref{fig:dynamic}.

As we can see, the performance by fixed experts is worse than DEEM with dynamic experts, even if we set the top-3 ones with the highest frequency or accuracy.
The performance by using the experts with the least frequency and accuracy further decreases the performance. These findings show that dynamically retrieving the suitably experienced experts according to specific sentences is useful.

\subsection{Impact of Demonstrations, Experts, and Discussion Turns}
For retrieving the experts and prompting LLMs for predictions  (Section~\ref{section:33}), we investigate the impact of the numbers of demonstrations, retrieved experts, and discussion turns. The overall results are shown in Figure~\ref{fig:numbers}.

\noindent\textbf{Demonstrations.} 
The results depicted in Figure~\ref{fig:numbers}(a) indicate sub-optimal performance during zero- or one-shot reasoning, i.e., when the number of demonstrates is 0 or 1. However, the performance progressively improves and eventually plateaus with the presence of 2 or more demonstrates. This trend suggests that LLMs can deliver commendable performance when provided with exemplars labeled both 'favor' and 'against'.

\noindent\textbf{Retrieved Experts.} 
In Figure~\ref{fig:numbers}(b), we find that 2 to 5 experts achieve the best performance, and the performance dramatically drops when the experts are 10 or more. This shows that engaging more experts does not consistently lead to better performance, which can be due to introducing noise from unrelated experts, as discussed in Section.~\ref{section:filtering_strategy}.

\noindent\textbf{Discussion Turns.}
From Figure~\ref{fig:numbers}(c), it is apparent that a single turn can already yield satisfactory performance. Additional turns (e.g., 3 or 4) bring improvement. This could be attributed to two primary factors: 1) The difficulty in generating high-quality expert discussions increases with multiple turns, potentially diluting the effectiveness of demonstrations. 2) Complex multi-step reasoning may not be a requisite for stance detection within a sentence.

\subsection{Effect of Expert Modeling}
To investigate the effect of ``experts'', we compare self-consistency reasoning and two variants of our proposed DEEM. The results are shown in Table~\ref{table:compare_sc}.

\begin{table}[t!]
	\centering
   \resizebox{\columnwidth}{!}{
	\begin{tabular}{lccc}
	    \toprule 
\textbf{Method}&\textbf{DT-HC}&\textbf{DT-TC}&\textbf{HC-BS}\\
        \midrule
        Few-Shot&76.6&78.2&72.8\\
        Few-Shot + SC (N=3)&76.3&80.1&76.7\\
        
        DEEM w/ ``Person A/B/C''&77.2&79.0&77.0\\
        DEEM w/ ``Expert A/B/C''&78.6&78.7&76.7\\
        \textbf{DEEM} (Ours) &\textbf{81.7}&\textbf{80.7}&\textbf{83.5}\\
        
    \bottomrule
	\end{tabular}}
	\caption{Results of few-shot reasoning, self-consistency (SC) reasoning, and variants of our DEEM method. By leveraging multiple experts, our DEEM method can outperforms self-consistency-based methods in a single response.}
	\label{table:compare_sc}
\end{table}

\noindent\textbf{Comparison with Self-Consistency Reasoning.}
Unlike methods without using experts, our method models three experts and generates the prediction according to their analysis, thus it integrates contextualized information from multiple reasoning paths, which is similar to the self-consistency reasoning method~\cite{self-consistency}. The difference is that self-consistency reasoning requires multiple API calls and multiple responses to get the prediction through voting.
From the results we can see that using self-consistency reasoning can improve the few-shot results, showing that the single-reply approach limits the performance. Our method with multi-expert can obtain comparable or better predictions through a single response.

\noindent\textbf{Comparison with Substitute Roles}. In our method, we specify the profession of the experts, i.e., social media experts, political experts, etc. To make a comparison, we remove the profession and use the substitute roles ``expert A/B/C'' or ``person A/B/C'' to involve the multi-role discussion.
From the results we can see that our method with specified experts achieves the best results, showing that expert modeling is useful for stance detection, which can offer more reliable results.

\subsection{Multi-Experts as Bias Reduction}

\begin{table}[t!]
	\centering
   \resizebox{\columnwidth}{!}{
	\begin{tabular}{lcccc}
	    \toprule 
\textbf{Method}&\textbf{Trump}&\textbf{Biden}&\textbf{Sanders}& \textbf{Avg.}\\
        \midrule
        BERTweet&47.6&53.6&53.6&51.6\\
    DQA &43.9&45.7&45.2&44.9\\
    StSQA &54.5&55.1&53.6&54.4\\
        \textbf{DEEM} (Ours)&\textbf{60.9}&\textbf{64.8}&\textbf{67.6}&\textbf{64.4}\\
        
    \bottomrule
	\end{tabular}}
	\caption{Comparison between methods for the samples in P-Stance dataset with the label ``neutral''. The results show that our DEEM method show less bias towards these neutral expressions.}
	\label{table:bias_analysis}
\end{table}

Given the training data distribution, LLMs have shown bias towards specific targets. As ~\citet{zhang2023investigating} demonstrates, ChatGPT tends to show bias towards certain topics with no evident stance, either through direct prompting or chain-of-thought reasoning.

We explore whether the incorporation of multiple experts can mitigate this stance bias. We focus on sentences labeled as `neutral' and compare predicted stances using different methods. The results are shown in Table~\ref{table:bias_analysis}. For neutral stance samples, the outcomes from DQA and StSQA sometimes fall below or match those from the fine-tuned BERTweet model. This suggests that large models do carry some stance bias, corroborating ~\citet{zhang2023investigating}'s findings. However, our multi-expert method consistently yields the best results for `neutral' cases, demonstrating the potential of multiple experts in reducing stance bias, even in the few-shot settings without further parameter updates.

\begin{figure}[t!]
    \centering
    \includegraphics[width=\columnwidth]{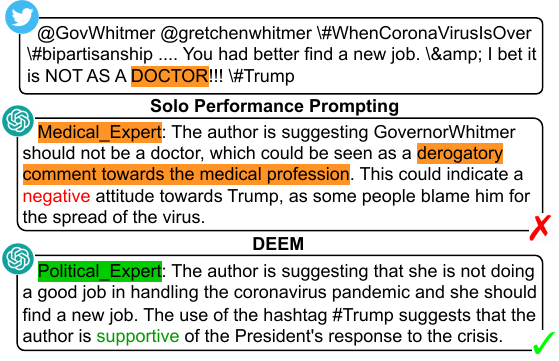}
    \caption{Example of solo performance prompting~\cite{wang2023unleashing} with generated experts (Middle) and our method with filtered experienced experts through retrieval (Bottom).}
    \label{fig:case-study}
\end{figure}

\subsection{Case Study}
One example is given in Figure~\ref{fig:case-study} to show the comparison between the fully generated expert and the retrieved experienced expert. We can see that LLMs misunderstand the text ``DOCTOR'' and generate the unmatched ``Medical Expert'', giving incorrect responses with hallucinations. This reflects an important issue with current LLMs in role-playing, which is the inability to generate appropriate characters to solve specific tasks when the roles are not provided, especially in non-general domain problems.

In contrast, our DEEM method can determine that this is not a problem related to the medical domain based on the experience of collecting diverse experts, retrieving the experienced ``Political Expert'' according to this specific question and show much more reliable responses in the final.

\section{Conclusion}

We propose DEEM, a dynamic experienced expert modeling method for stance detection. Different from existing multi-agent reasoning methods, DEEM first generates possible diverse experts without leveraging domain knowledge and detailed expert descriptions, then filters the experienced experts to fulfill the generalizability and reliability, and finally it involves a retrieval mechanism during reasoning. Experimental results on both InstructGPT and ChatGPT model show that DEEM achieves a consistent improvement over all baselines across three benchmark datasets, outperforming methods with self-consistency reasoning, reducing the bias of LLMs, and mitigate potential hallucinations brought from generated inappropriate experts.
\section*{Limitations}
There are several limitations that should be considered for our proposed DEEM method:

\noindent \textbf{Limited Generalizability to Other Domains:}
The present study confines its scope to three established benchmark datasets, raising questions about the extent to which our DEEM method generalizes to diverse domains. A more comprehensive investigation into the method's applicability across various content categories and targets is warranted.

\noindent \textbf{Dependency on Large Language Models:}
The efficacy of DEEM heavily relies on the availability and accessibility of large language models. This dependence may pose challenges in resource-constrained environments, thus limiting the practical deployment of the proposed approach.

\bibliography{anthology, custom}
\bibliographystyle{acl_natbib}

\end{document}